\pdfoutput=1

\documentclass[11pt]{article}

\usepackage[]{emnlp2021}

\usepackage{times}
\usepackage{latexsym}

\usepackage[T1]{fontenc}

\usepackage[utf8]{inputenc}

\usepackage{microtype}

\usepackage{subcaption}
\usepackage{caption}
\usepackage{graphicx}

\usepackage{covington}

\usepackage{makecell}

\usepackage{booktabs}
\usepackage{multirow}
\usepackage{multicol}
\usepackage{comment}
\usepackage{eurosym}
\usepackage{csquotes}
\usepackage{amsthm,amsmath}
\usepackage{paralist}
\usepackage{xcolor}
\usepackage{fdsymbol}
\usepackage{epstopdf}

\title{DWUG: A large Resource of Diachronic Word Usage Graphs \\in Four Languages}

\author{Dominik Schlechtweg,$^{\clubsuit}$  Nina Tahmasebi,$^{\spadesuit}$ Simon Hengchen,$^{\spadesuit}$\\ \textbf{Haim Dubossarsky,$^{\vardiamondsuit}$ Barbara McGillivray$^{\diamondsuit,\heartsuit}$} \\
{{\tt semeval2020lexicalsemanticchange@turing.ac.uk}} \\
$^{\clubsuit}$University of Stuttgart, $^{\spadesuit}$University of Gothenburg, $^{\vardiamondsuit}$University of Cambridge,\\ $^{\diamondsuit}$King's College London $^{\heartsuit}$The Alan Turing Institute
}

\begin{document}
\maketitle
\begin{abstract}

Word meaning is notoriously difficult to capture, both synchronically and diachronically. In this paper, we describe the creation of the largest resource of graded contextualized, diachronic word meaning annotation in four different languages, based on 100,000 human semantic proximity judgments. We describe in detail the multi-round incremental annotation process, the choice for a clustering algorithm to group usages into senses, and possible -- diachronic and synchronic -- uses for this dataset. 
\end{abstract}

\section{Introduction}

The view on word meaning and senses in computational linguistics has moved from a \textbf{discrete} \citep{Weaver49,Navigli09} to a \textbf{graded} \citep{McCarthyNavigli2009,Erk09investigationson,Erk13,Schlechtwegetal18} perspective. However, scalable annotation strategies for this graded view yielding large-scale data for semantic evaluation have not been implemented yet. We build on two pre-existing schemata for graded contextual word meaning annotation \citep{Erk13} and show how they can be applied efficiently to create large-scale data in a diachronic setup. 

Both procedures populate a \textbf{Word Usage Graph} \citep[WUG,][]{carthy16,schlechtweg-etal-2020-semeval} for a target word with annotator judgments. Procedure (i) requires annotators to judge usage pairs on a semantic proximity scale avoiding the a priori definition of word senses. This makes it preparation-lean and reduces experimenter influence. Procedure (ii) relies on a predefined list of senses and requires annotators to judge usage-sense pairs on the same proximity scale as in procedure (i). Both procedures avoid binary assignments of word senses to word usages, which have been shown to be inadequate in many cases \citep{Kilgarriff1997,Hanks2000,Kilgarriff2006}. The resulting graphs relate word usages to each other (either directly or indirectly) and thus allow for a posteriori hard- or soft-clustering, where clusters can be interpreted as senses \citep{Schutze1998,carthy16,schlechtweg-etal-2020-semeval}. This makes the collapsing of senses possible, while allowing for sense overlap where this seems adequate \textit{after} observing the annotated data. While both procedures require more judgments than traditional discrete word sense annotation, we show how the sampling of word usages can be optimized to reduce the number of necessary judgments.

We apply the above-described annotation procedures in a multi-lingual diachronic setup to create Diachronic WUGs (DWUGs). These contain annotations of the usages of a set of target words in corpora from two time periods \citep{schlechtweg-etal-2020-semeval}. This allows us to identify changes in the WUGs over time. The final resource contains 168
DWUGs for four different languages (English (EN), German (DE), Swedish (SV), Latin (LA)) relying on approximately 100,000 human judgments.\footnote{We provide DWUGs as Python NetworkX graphs, the raw  annotated data, descriptive statistics, inferred clusterings, change values and interactive  visualizations at \url{https://www.ims.uni-stuttgart.de/data/wugs}.}

After describing the annotation procedure, we provide a detailed analysis of annotator disagreements and evaluate the robustness of the annotated graphs. DWUGs can be exploited in many ways:
\begin{itemize}
 \setlength{\itemsep}{0pt}
\setlength{\parskip}{0pt}
\setlength{\parsep}{0pt}
    \item as large sets (thousands) of pairwise semantic proximity judgments to evaluate contextualized embeddings in multiple languages;
    \item the inferred change scores can be used to evaluate semantic change detection models
    \item as word sense disambiguation/discrimination resources with additional aspects such as variation over time;
    \item the graphs may be treated as research objects in their own right, providing insights on cognitive aspects of word meaning and posing practical problems such as finding robust and efficient clustering algorithms.
\end{itemize}

\section{Related Work}

There has been a significant shift in the view on word meaning and word senses in computational linguistics since the birth of the field. The early formulations of the Word Sense Disambiguation (WSD) task took a \textbf{discrete} view on word senses, assuming a fixed inventory of senses and a single best sense per word usage \citep{Weaver49,Navigli09}. After this view was shown empirically to be inadequate \citep{Kilgarriff1997,Hanks2000,Kilgarriff2006}, researchers have increasingly adopted a \textbf{graded} view on word senses, whereby a word usage may be assigned to multiple senses and more fine-grained distinctions are allowed within senses \citep{McCarthyNavigli2009,Erk09investigationson,Erk13}. 

Moreover, various approaches on how senses can be qualified have been proposed, starting from manual sense descriptions \citep{Wilks75}, to representing a sense solely by clusters of word usages \citep{Schutze1998} or by lexical substitutes \citep{McCarthyNavigli2009}. Recently, developments on computational models of the meaning of individual word usages \citep{peters-etal-2018-deep,devlin-etal-2019-bert} have inspired new research on graded word meaning \citep{Armendariz19}.

For discrete word senses, large-scale annotation projects have been carried out, e.g. SemCor and OntoNotes \citep{Langoneetal04,OntoNotes2006}. An advantage of the graded approach is that, through bypassing sense definitions, major parts of the annotation pipeline can be automated \citep[cf.][]{Biemann-2013-system}. Studies on graded word meaning, however, cover only small amounts of data \citep{SoaresdaSilva1992,Brown2008,McCarthyNavigli2009,Erk09investigationson,Erk13,haettySurel-2019}. 

The above-mentioned studies have paved the way to study diachronic dimensions of meaning. So far, studies that have explicitly tried to capture this dimension are rare, small-scale and mostly assume discrete word senses \citep{Bamman11p1,Lau12p591,Cook14p1624,Tahmasebi17,schlechtweg-EtAl-2017-CoNLL,Perrone19,diacrita_evalita2020,perrone2021latin-greek}. The most recent approaches take a graded view \citep{giulianelli-etal-2020-analysing,rodina2020rusemshift} building on the DURel framework \citep{Schlechtwegetal18}, but result in little annotated data. We release the largest known resource of diachronic contextualized graded word meaning. Our resource is related to discrete word sense annotation resources such as SemCor or OntoNotes in providing groups of word usages with the same/similar senses. However, they differ from those resources in the way in which senses are obtained, i.e., inferred on the pairwise annotated data and the graded nature of usage-usage and usage-sense comparisons. In this, our resources are strongly related to USim and WSim-2 \citep{Erk13}, but differ from these by the additional diachronic dimension, the size of the graphs and the principled and robust approach to clustering.

\begin{table}[t]
\small
\centering
\begin{tabular}{lll}
\toprule
        & $C_1$ & $C_2$ \\
\midrule
\textbf{English}  & CCOHA 1810--1860  &  CCOHA 1960--2010 \\
\textbf{German}  & DTA 1800--1899  & BZ+ND 1946--1990 \\
\textbf{Swedish}  & Kubhist 1790--1830 & Kubhist 1895--1903 \\
\textbf{Latin}  & LatinISE -200--0 & LatinISE 0--2000 \\
\bottomrule
\end{tabular}
\caption{Time-defined subcorpora for each language from which annotation data was sampled.}\label{tab:corpora}
\vspace{-8ex}
\end{table}

\section{Data}
\label{sec:data}

The data for annotation was sampled from two time-specific historical subcorpora for each language as summarized in Table \ref{tab:corpora}.
For English, we used the Clean Corpus of Historical American English \cite[CCOHA,][]{davies2002corpus,alatrash-etal-2020-ccoha}, which spans 1810s--2000s.\footnote{Additional pre-processing steps were needed for English: for copyright reasons CCOHA contains frequent replacement tokens (10 x `@'). We split sentences around replacement tokens and removed them.} For German, we used the DTA corpus \cite{dta2017} and a combination of the BZ and ND corpora \cite{BZ2018,ND2018}. DTA contains texts from different genres spanning the 16th--20th centuries. BZ and ND are newspaper corpora jointly spanning 1945--1993. For Latin, we used the LatinISE corpus \cite{mcgillivray-kilgarriff} spanning from the 2nd century B.C. to the 21st century A.D.\footnote{LatinISE is automatically lemmatised and part-of-speech tagged.
A study on lemmatisation accuracy on a sample of two texts (Cicero's \emph{De Officiis} and Rutilius Taurus Aemilianus Palladius' \emph{Opus agriculturae} against the PROIEL treebank as a gold standard \cite{haugjohndal} (\url{https://proiel.github.io/}).) showed an accuracy of 92.77\% and 80.96\%, respectively.} For Swedish, we used the Kubhist corpus \cite{KubHist}, a newspaper corpus containing texts from 18th--20th century. The corpora are automatically lemmatised and POS-tagged. CCOHA and DTA are spelling-normalized. BZ, ND and Kubhist contain frequent OCR errors \cite{adesam2019exploring,hengchen2020vocab}.

For each language half of the target words ($\approx$ 20) were chosen as words for which a change between $C_1$ and $C_2$ was described in etymological or historical dictionaries \cite{oxford2009oxford,Paul02XXI,clackson,svenska}. The other half was determined by sampling a control counterpart with the same POS and comparable frequency development between $C_1$ and $C_2$ as the corresponding target word. (For details refer to \citet{schlechtweg-etal-2020-semeval}.)

\begin{table}[t]
\parbox{.45\linewidth}{
\centering
\tabcolsep=0.11cm
\begin{tabular}{ll}
\multirow{4}{*}{$\Bigg\uparrow$}&Identity\\
&Context Variance\\
&Polysemy\\
&Homonymy
\end{tabular}
\label{tab:blank}}
\hfill
\parbox{.53\linewidth}{
\centering
\begin{tabular}{ll}
\multirow{4}{*}{$\Bigg\uparrow$} &4: Identical\\
 &3: Closely Related\\
 &2: Distantly Related\\
 &1: Unrelated\\
\end{tabular}
\label{tab:scale2}}
\caption{\citet{Blank97XVI}'s continuum of semantic proximity (left) and the DURel relatedness scale derived from it (right).}\label{tab:scales}
\vspace{-8ex}
\end{table}

\begin{figure*}[t]
    \begin{subfigure}{0.33\textwidth}
\frame {
\includegraphics[width=\linewidth]{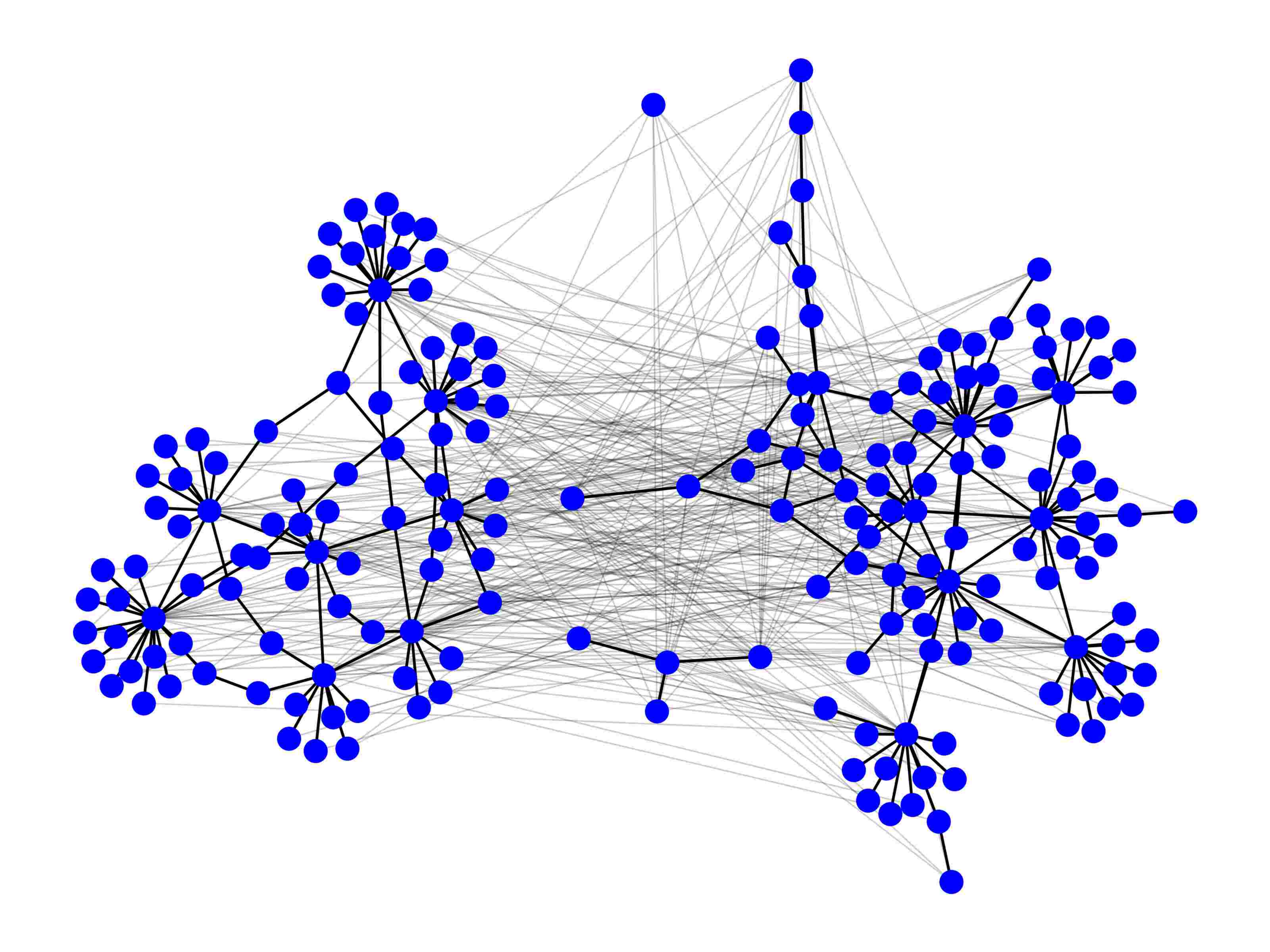}
}
       
    \end{subfigure}
    \begin{subfigure}{0.33\textwidth}
\frame {        \includegraphics[width=\linewidth]{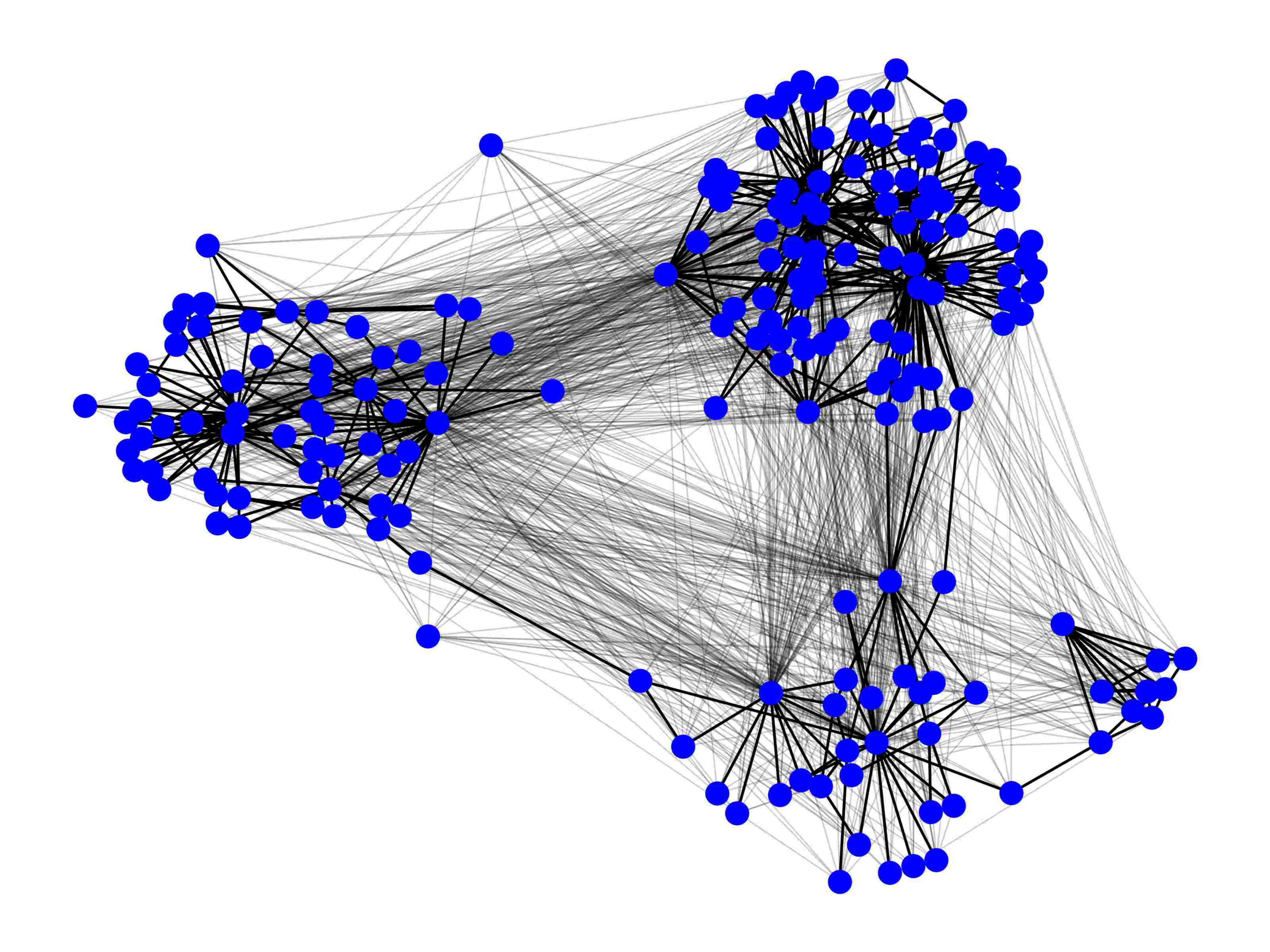}}
       
    \end{subfigure}
    \begin{subfigure}{0.33\textwidth}
\frame {        \includegraphics[width=\linewidth]{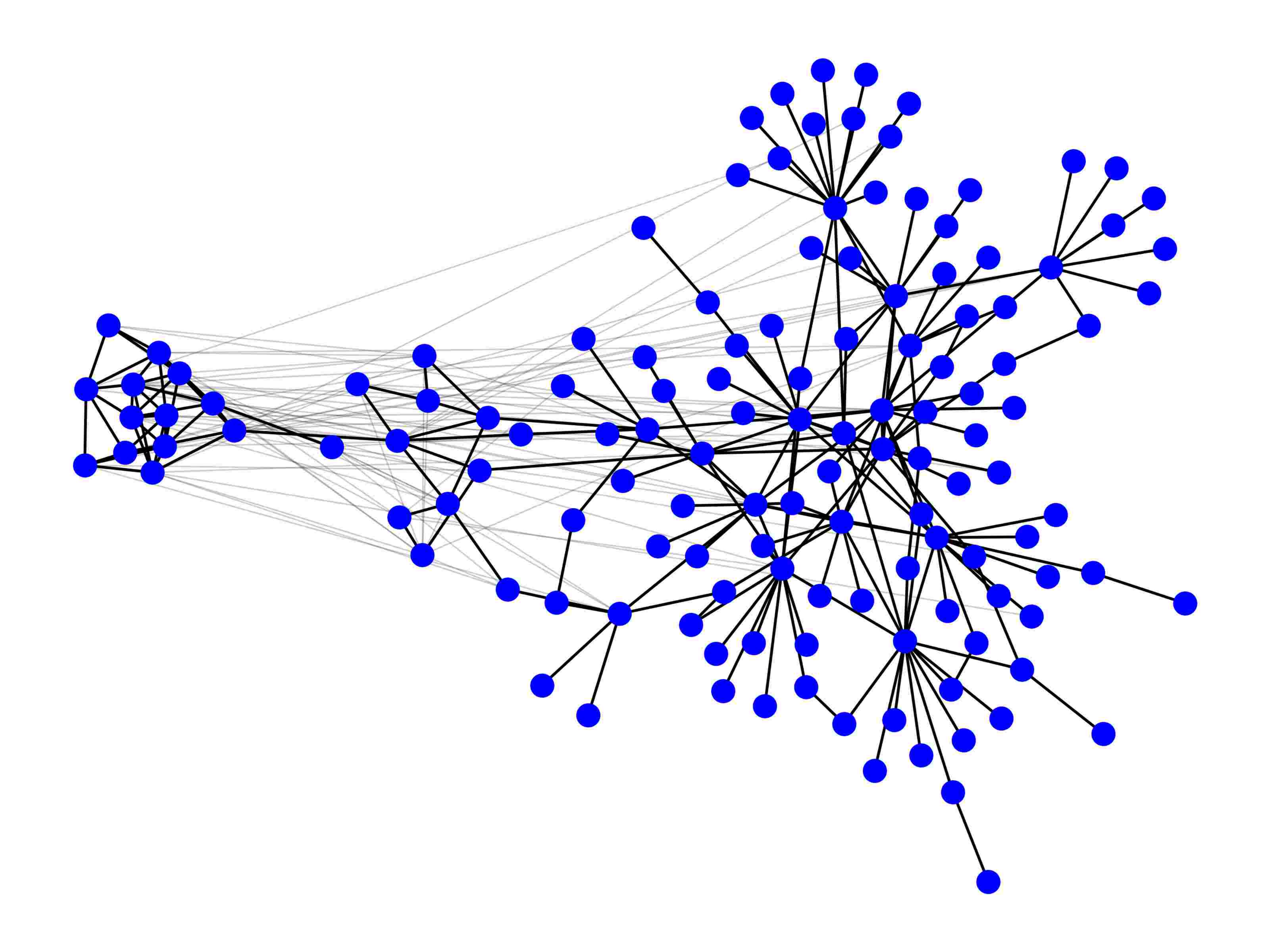}}
       
    \end{subfigure}
    \caption{Usage-usage graphs of English \textit{plane} (left), German \textit{ausspannen} (middle) and Swedish \textit{ledning} (right). Nodes represent usages of the respective target word. Edge weights represent the median of relatedness judgments between usages (\textbf{black}/\textcolor{gray}{gray} lines for \textbf{high}/\textcolor{gray}{low} edge weights, i.e., weights $\geq$ 2.5/weights $<$ 2.5).}\label{fig:triplet1}
\end{figure*}

\section{Procedure (i): Usage-Usage Graphs}
\label{sec:procedure1}

We first describe the procedure devised to annotate EN, DE and SV data and later describe the procedure for LA in Sec.~\ref{sec:procedure2}. A usage-usage graph (UUG) $\mathbf{G = (U, E, W)}$ is a weighted, undirected graph, where nodes $u \in U$ represent word usages and weights $w \in W$ represent the semantic proximity of a pair of usages (an edge) $(u_1,u_2) \in E$ \citep{carthy16,schlechtweg-etal-2020-semeval}. 
In practice, semantic proximity can be measured by human annotator judgments on a scale of relatedness \citep{Brown2008,Schlechtwegetal18} or similarity \citep{Erk13}. The annotation procedure starts from a non-annotated sample of word usages and aims to populate a UUG for each target word in several rounds of annotation with human judgments of semantic relatedness.\footnote{A similar annotation procedure is implemented in the openly accessible DURel annotation interface: \url{https://www.ims.uni-stuttgart.de/data/durel-tool}.}  Annotators were asked to judge the semantic relatedness of pairs of word usages using the scale in Table \ref{tab:scales}. (\ref{ex:1}) and (\ref{ex:2}) show two example usages of the noun \textit{plane}.
\begin{example}\label{ex:1}
Von Hassel replied that he had such faith in the \textbf{plane} that he had no hesitation about allowing his only son to become a Starfighter pilot.
\end{example}%
\begin{example}\label{ex:2}
This point, where the rays pass through the perspective \textbf{plane}, is called the seat of their representation.
\end{example}
Figure \ref{fig:triplet1} shows three UUGs resulting from our annotation.

\subsection{Annotators}
We started out with four annotators per language. Following high annotation loads and dropouts, additional annotators were hired, resulting in 9/8/5 total annotators for EN/DE/SV, respectively. All annotators were native speakers and current or former university students. The number of annotators with a background in historical linguistics was two for DE and one for EN and SV.\footnote{\citet{Schlechtwegetal18} observe that annotators with and without historical background have high agreement.}

\subsection{Usage sampling} 
\label{sec:use1}

We refer to an occurrence of a word $w$ in a sentence by `usage of $w$'. For each target word, 100 usages were randomly sampled from each of $C_1$ and $C_2$ (Table \ref{tab:corpora}). Each usage contained the target word in its lemma form and a minimum of ten tokens, yielding a total of 200 usages per target word.\footnote{Because English frequently combines various POS in one lemma and many of our target words underwent POS-specific semantic changes, we sampled only usages of English target words with the broad POS tag for which a change had been described.} If a target word had less than 100 usages, the full sample was annotated. The usage samples were subsequently mixed into a joint set $U$ per target word. The set of usages $U$ were annotated by presenting usage pairs to annotators in randomized order, hence, the annotators did not know from which time period each usage stemmed.

\subsection{Edge sampling} 
\label{sec:edge1}

Annotating the full usage graph is not feasible even for a small set of $n$ usages as this implies annotating $n*(n-1)/2$ edges. Hence, the main challenge with this annotation approach was to annotate as few edges as possible, while keeping the information needed to infer a meaningful clustering on the graph. This was achieved by annotating the data in several rounds. After each round, the UUG of a target word was updated with the new annotations and a new clustering was obtained.\footnote{If an edge was annotated by several annotators, the median was retained as an edge weight.} Based on this clustering, the edges for the next round were sampled through heuristics similar to \citet{Biemann-2013-system}. The annotation load was randomly distributed
making sure that roughly half of the usage pairs were annotated by more than one annotator. 

The first round aimed to obtain a small high-quality reference set of clusters. This was achieved through the sampling of 10\% of the usages from $U$ and 30\% of the edges by a random walk through the sample graph (\textbf{exploration}), which guaranteed that all nodes are connected by some path. Hence, the first clustering was obtained on a small but richly-connected subgraph ensuring that not too many clusters were inferred, as this would lead to a strong increase in annotation instances in the subsequent rounds. In the second round, the reference clusters from the first round served as a comparison for those usages which were not assigned to a multi-cluster yet (\textbf{combination}).\footnote{We refer to a cluster with $\geq 2$ usages as `multi-cluster'.} In all subsequent rounds, both a combination step and an exploration step were employed. The combination step combined each single usage $u_1$ which is not yet member of a multi-cluster with a random usage $u_2$ from each of the multi-clusters to which $u_1$ had not yet been compared. The exploration step consisted of a random walk on 30\% of the edges from the non-assignable usages, i.e., usages which had already been compared to each of the multi-clusters but were not assigned to any of these by the clustering algorithm. This procedure slowly populated the graph while \textit{minimizing the annotation of redundant information}.
We aimed to stop the procedure when each cluster had been compared to each other cluster. The sample sizes for the random walk were tuned and validated in a simulation study \citep{schlechtweg-etal-2020-semeval}.

\begin{figure*}[t]
    \begin{subfigure}{0.33\textwidth}
\frame {        \includegraphics[width=\linewidth]{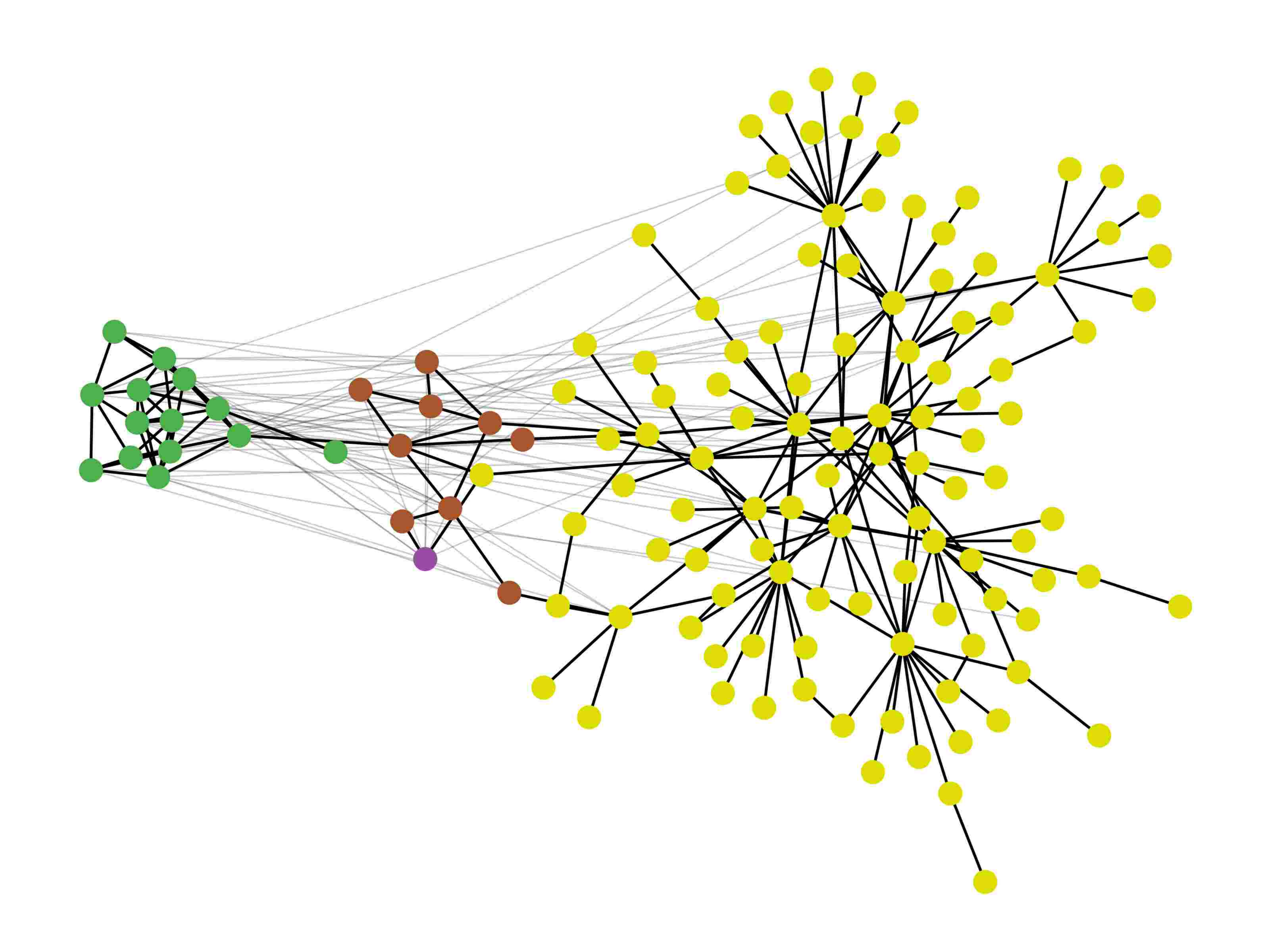}}
        \caption*{$G$}
    \end{subfigure}
    \begin{subfigure}{0.33\textwidth}
\frame{        \includegraphics[width=\linewidth]{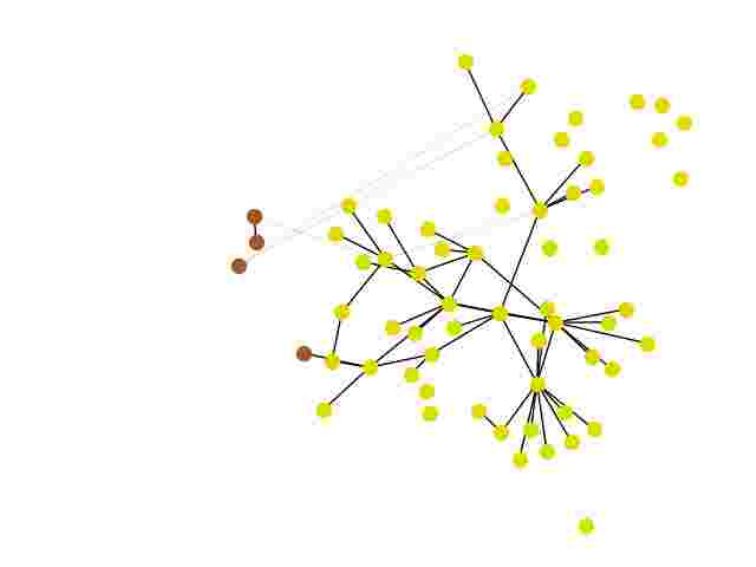}}
        \caption*{$G_1$}
    \end{subfigure}
    \begin{subfigure}{0.33\textwidth}
\frame{        \includegraphics[width=\linewidth]{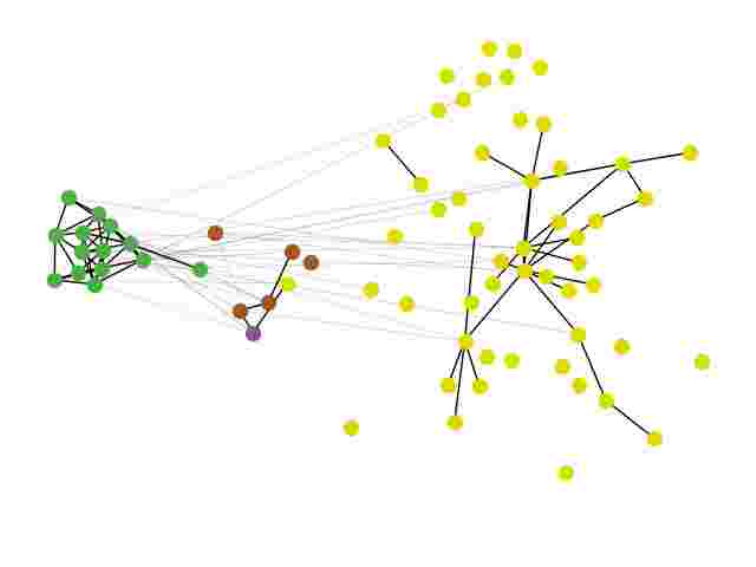}}
        \caption*{$G_2$}
    \end{subfigure}
    \caption{Usage-usage graph of Swedish \textit{ledning} (left), subgraph for first time period $C_1$ (middle) and second time period $C_2$ (right).}\label{fig:graphledning}
\end{figure*}

\begin{inparaenum}[(i)]
The above procedure was combined with further heuristics added after round 1 to increase the quality of the annotation: \item sampling a low number of randomly chosen edges and edges between already confirmed multi-clusters for further annotation to corroborate the inferred structure; \item detecting relevant disagreements between annotators, i.e., judgments with a difference of $\geq 2$ on the scale or edges with a median $\approx 2.5$, and redistributing the corresponding edges to another randomly chosen annotator from the ones who did not annotate the respective edge yet to resolve the disagreements; and \item detecting clustering conflicts, i.e., positive edges between clusters and negative edges within clusters (see below) and sampling a new edge for each node connected by a conflicting edge. This added more information in regions of the graph where finding a good clustering was hard. Furthermore, after each round, nodes from the graph whose 0-judgments (undecidable) made up more than half of their total judgments were removed, and in a few cases, whole words were removed if they had a high number of `0' judgments or needed a high number of further edges to be annotated. The annotation was stopped after four rounds for time constraints.
\end{inparaenum}
(An example of our annotation pipeline can be found in Appendix \ref{sec:pipeline}.)

\subsection{Clustering}
\label{sec:clustering1}

Tasks such as SemEval-2020 Task 1 require to derive a hard-clustering from the graphs.\footnote{However,  they also allow for soft-clustering reflecting the gradedness of word senses, which is an avenue for future work using this resource.}
The UUGs obtained from the annotation were weighted, undirected, sparsely observed and noisy. This called for a robust clustering algorithm. For this, a variation of correlation clustering \citep{Bansal04,schlechtweg-etal-2020-semeval} was employed minimizing the sum of cluster disagreements, i.e., the sum of negative edge weights within clusters plus the sum of positive edge weights across clusters. To see this, consider \citet{Blank97XVI}'s continuum of semantic proximity and the DURel relatedness scale derived from it, as illustrated in Table \ref{tab:scales}. In line with Blank, usage pairs with judgments of 3 and 4 are expected to belong to the same sense, while judgments of 1 and 2 belong to different senses. 
Consequently, the weight $W(e)$ of all edges $e \in E$ in a UUG $\mathbf{G = (U, E, W)}$ are shifted to $W'(e)=W(e)-2.5$ (e.g. a weight of $4$ becomes $1.5$). Those edges $e$ with a weight $W'(e) \geq 0$ are referred to as \textbf{positive} edges $P_E$, while edges with weights $W'(e)<0$ are called \textbf{negative} edges $N_E$. 
Let further $C$ be some clustering on $U$, $\phi_{E,C}$ be the set of positive edges \textbf{across} any of the clusters in clustering $C$ and $\psi_{E,C}$ the set of negative edges \textbf{within} any of the clusters. 
We then search for a clustering $C$ that minimizes $L(C)$:

\begin{equation}\label{eq:loss}
L(C) = \sum_{e\in \phi_{E,C}} W'(e) + \sum_{e\in \psi_{E,C}} |W'(e)| ~~.
\end{equation}
That is, the sum of positive edge weights between clusters and (absolute) negative edge weights within clusters is minimized. Minimizing $L$ is a discrete optimization problem which is NP-hard \cite{Bansal04}, which is eased by the relatively low number of nodes ($\leq 200$). Hence, the global optimum can be approximated sufficiently with a standard optimization algorithm such as Simulated Annealing \cite{Pincus1970}: an algorithm that has shown superior performance in a  previous simulation study by  \citet{schlechtweg-etal-2020-semeval}. Since we do not have strong efficiency constraints, we follow the same procedure.
In order to reduce the search space, we iterate over different values for the maximum number of clusters. We also iterate over randomly, as well as heuristically, chosen initial clustering states.\footnote{We used mlrose to perform the clustering \cite{Hayes19}. Find our code at \url{https://www.ims.uni-stuttgart.de/data/wugs}.}
This way of clustering usage graphs has several advantages: (i) It finds the optimal number of clusters on its own. (ii) It easily handles missing information (non-observed edges). (iii) It is robust to errors by using the global information on the graph. That is, one wrong judgment can be outweighed by correct ones. (iv) It directly optimizes an intuitive quality criterion on usage graphs. Many other clustering algorithms such as Chinese Whispers \cite{Biemann2006} make local decisions, so that the final solution is not guaranteed to optimize a global criterion such as $L$. (v) By weighing each edge with its (shifted) weight, $L$ respects the gradedness of word meaning. That is, edges with $|W'(e)| \approx 0$ have less influence on $L$ than edges with $|W'(e)| \approx 1.5$.
The clustered graphs are provided with the published data.
Figure \ref{fig:graphledning} ($G$) shows the annotated and clustered UUG for SV \textit{ledning}. Nodes represent usages of the target word (isolates removed). Edges represent the median of relatedness judgments between usages. Colors make clusters (senses) inferred on the full graph $G$. $G_1$ (left) and $G_2$ (right) represent the time-specific subgraphs resulting from removing the respective nodes and their edges for each time period ($C_1$, $C_2$) from the full graph.

\begin{figure*}[t]
    \begin{subfigure}{0.33\textwidth}
\frame{        \includegraphics[width=\linewidth]{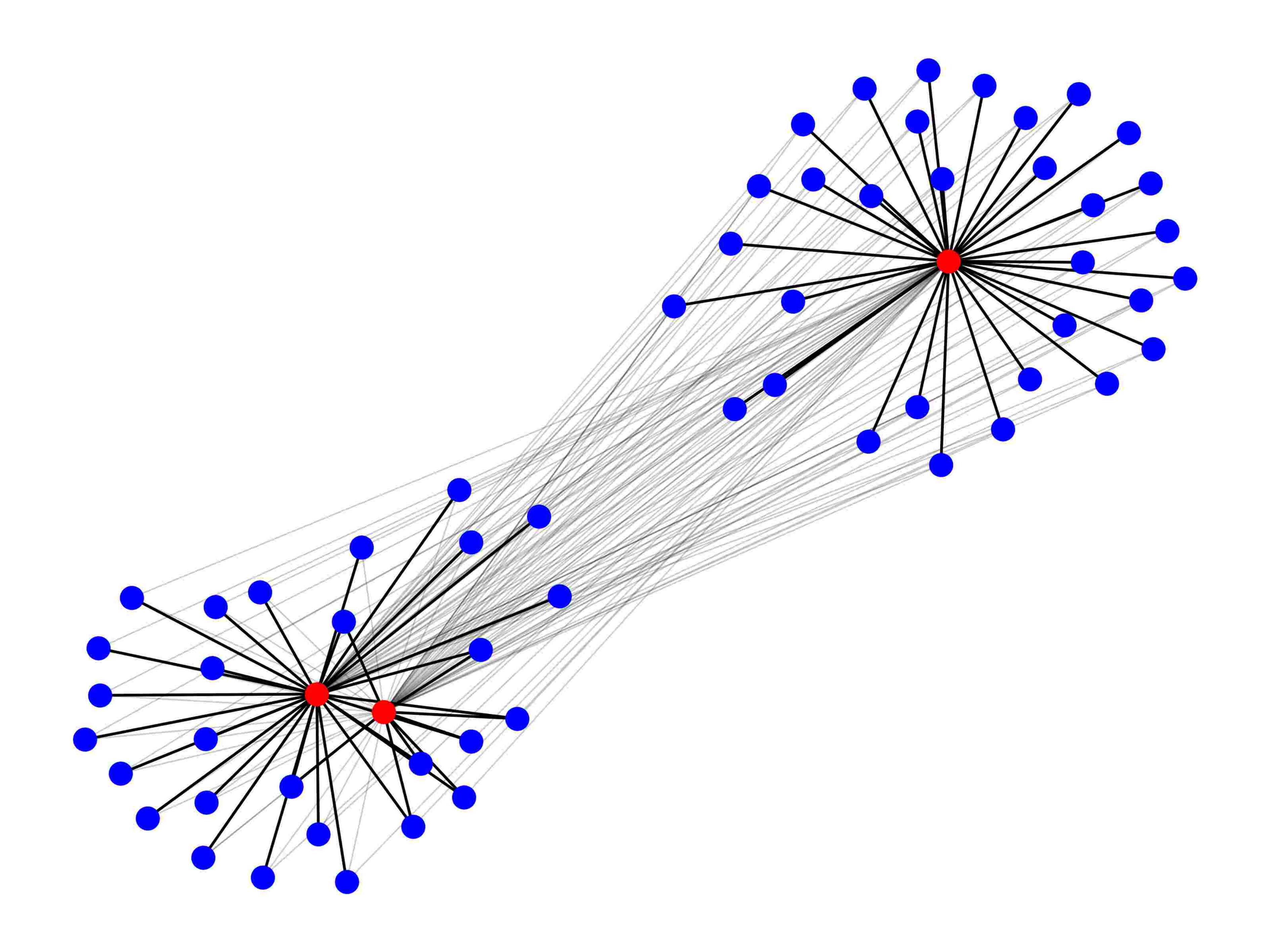}}
       
    \end{subfigure}
    \begin{subfigure}{0.33\textwidth}
\frame {        \includegraphics[width=\linewidth]{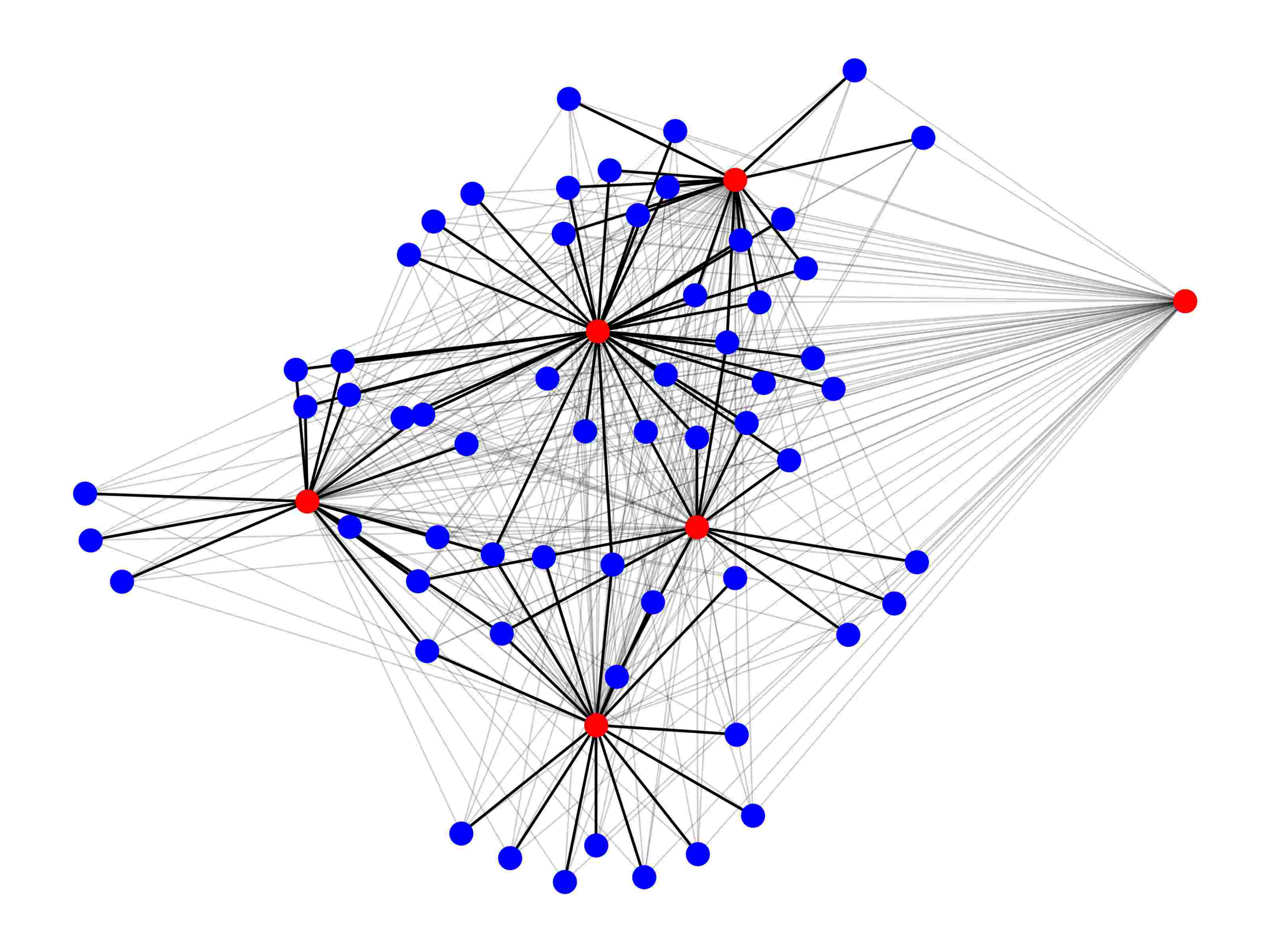}}
       
    \end{subfigure}
    \begin{subfigure}{0.33\textwidth}
\frame {        \includegraphics[width=\linewidth]{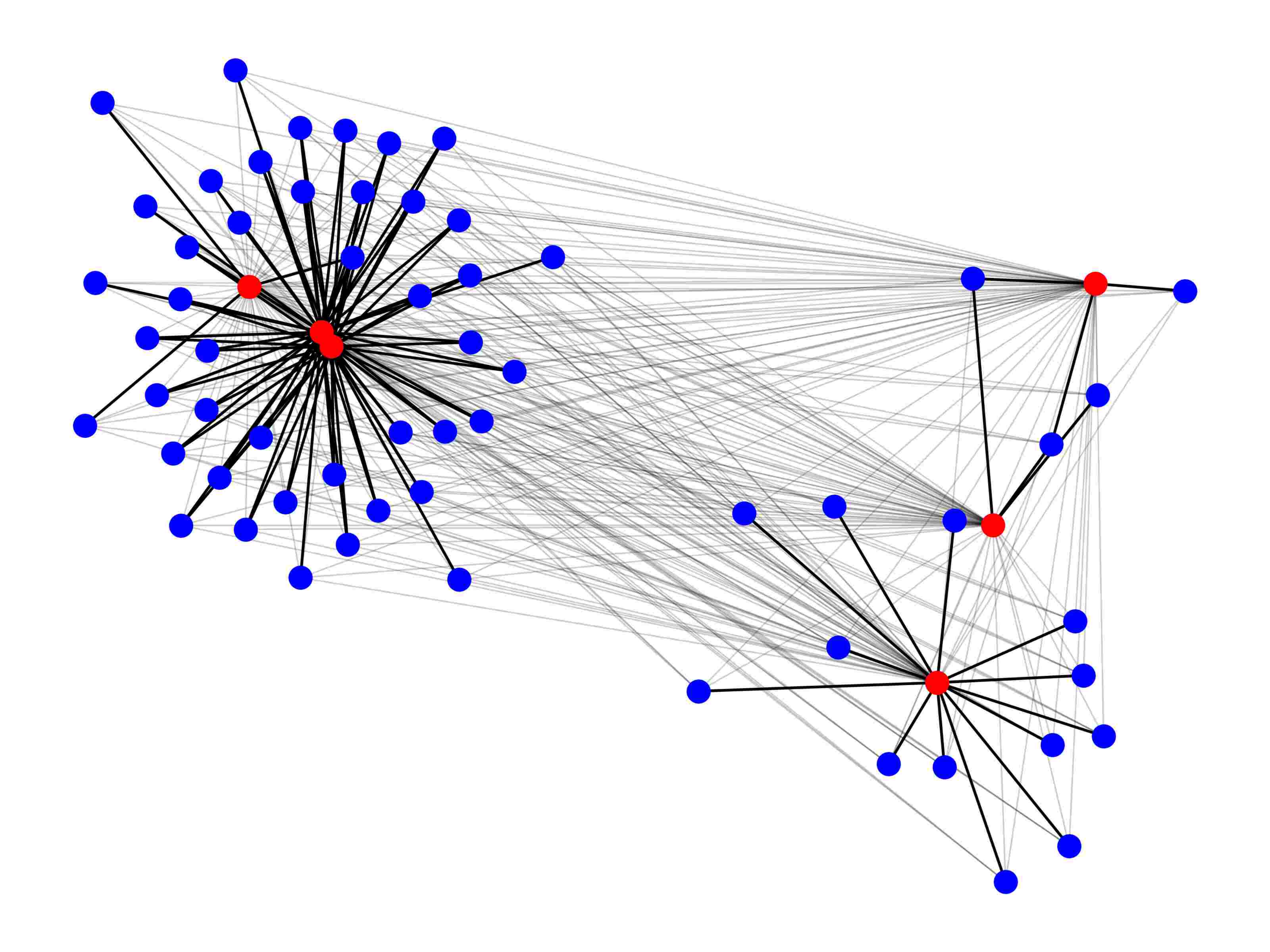}}
       
    \end{subfigure}
    \caption{Usage-sense graphs of Latin \textit{pontifex} (left), \textit{potestas} (middle) and \textit{sacramentum} (right). Nodes in \textcolor{blue}{blue}/\textcolor{red}{red} represent \textcolor{blue}{usages}/\textcolor{red}{senses} respectively.}\label{fig:triplet2}
\end{figure*}

\section{Procedure (ii): Usage-Sense Graphs}
\label{sec:procedure2}

In this section, we describe the procedure devised to annotate the Latin data. This procedure is different from the other languages, as in a trial annotation task the annotators reported difficulties to judge usage-usage pairs. In consideration of this, usage-sense graphs were employed. Since we do not have access to native speakers of Latin, eight annotators with a high-level knowledge of Latin were recruited, ranging from undergraduate students to PhD students, post-doctoral researchers, and more senior researchers.

\subsection{Usage-sense graphs}
A usage-sense graph (USG) $\mathbf{G = (V, E, W)}$ is a weighted, undirected graph, whose nodes $v \in V$ represent either word usages or sense descriptions and weights $w \in W$ represent the semantic proximity of a usage-sense pair $(u_1,s_1) \in E$.\footnote{Note that we do not consider the possible cases where  $E$ contains additional usage-usage pairs or sense-sense pairs.} We denote the set of word usages as $U$ and the set of word sense descriptions as $S$, where $V = U \cup S$. Following \citet{Erk13}, semantic proximity can be measured by human annotator judgments on a similar scale as for USGs. Hence, we started from a non-annotated sample of usage-sense pairs and populated a USG for each target word with human judgments of semantic relatedness. Annotators were asked to judge the semantic relatedness of usage-sense pairs using the scale as for the other languages.
(\ref{ex:3})
contains an example of a usage-sense pair for \textit{sacramentum}, displaying the older sense ``a civil suit or process''
\begin{example}\label{ex:3}
\texttt{Usage:} Cum Arretinae mulieris libertatem defenderem et Cotta xviris religionem iniecisset non posse nostrum \textbf{sacramentum} iustum iudicari, [\ldots
\\
{\em `When I was defending the liberty of a woman of Arretium, and when Cotta had suggested a scruple to the decemvirs that our \textbf{action} was not a regular one, [\ldots]
'}\footnote{M. Tullius Cicero. The Orations of Marcus Tullius Cicero, literally translated by C. D. Yonge, B. A. London. Henry G. Bohn, York Street, Covent Garden. 1856.}\\
\texttt{Sense:} ``a cause, a civil suit or process''
\end{example}%
Figure \ref{fig:triplet2} shows three USGs resulting from our annotation. The first word, \emph{pontifex}, originally meant ``a member of the college of priests having supreme control in matters of public religion in Rome'', and with Christianity it acquired the sense of ``bishop''. The three senses presented to the annotators were ``priest, high priest'', ``Roman high-priest, a pontiff, pontifex'', and ``bishop''. The first two correspond to the two red nodes in the bottom left corner of the first plot in Figure \ref{fig:triplet2}, and the last one corresponds to the top right red node. The plot of the second word, \emph{potestas} shows the complex and highly related set of its senses, which can be summarised as: ``Power of doing any thing''; ``Political power''; ``Magisterial power''; ``Meaning of a word'' (the isolated sense on the far right of the plot); ``Force, efficacy''; ``Angelic powers''. The last plot refers to \emph{sacramentum} and shows how the two senses ``military oath of allegiance'' and ``oath'' are close together on the top left of the plot, while the legal sense ``a civil suit or process'' is separated from the others in the top right corner and the Christian sense of ``sacrament'' is at the bottom right corner.

\begin{figure*}[t]
    \begin{subfigure}{0.33\textwidth}
\frame {        \includegraphics[width=\linewidth]{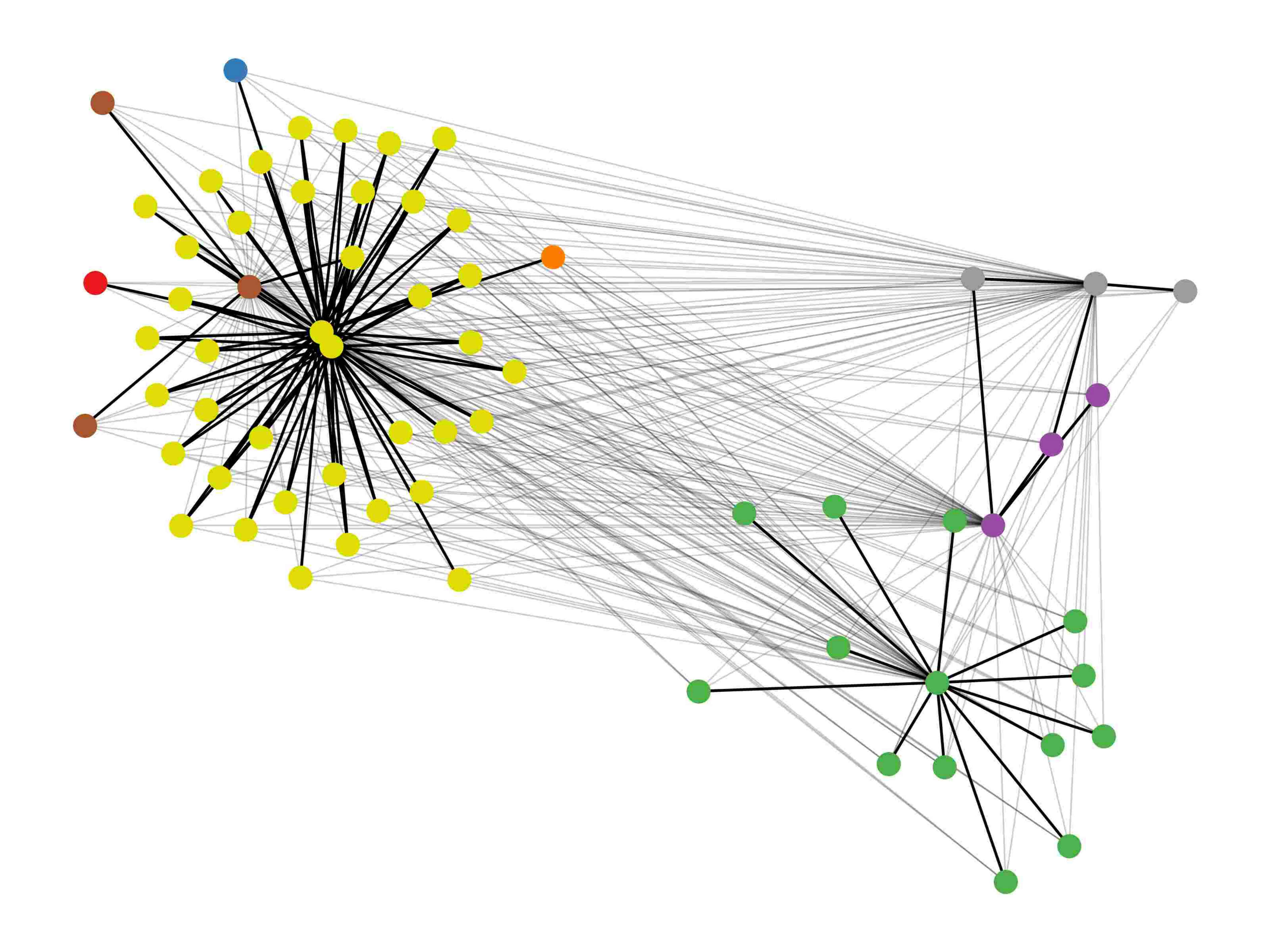}}
        \caption*{$G$}
    \end{subfigure}
    \begin{subfigure}{0.33\textwidth}
\frame{        \includegraphics[width=\linewidth]{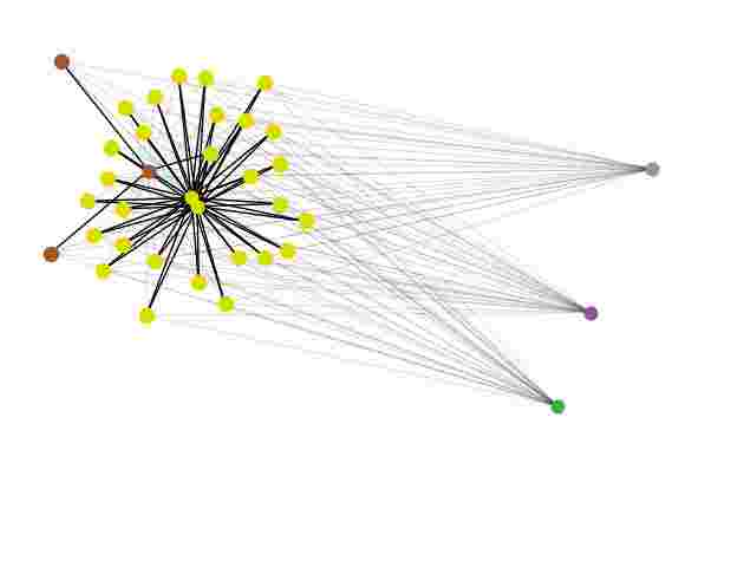}}
        \caption*{$G_1$}
    \end{subfigure}
    \begin{subfigure}{0.33\textwidth}
\frame{        \includegraphics[width=\linewidth]{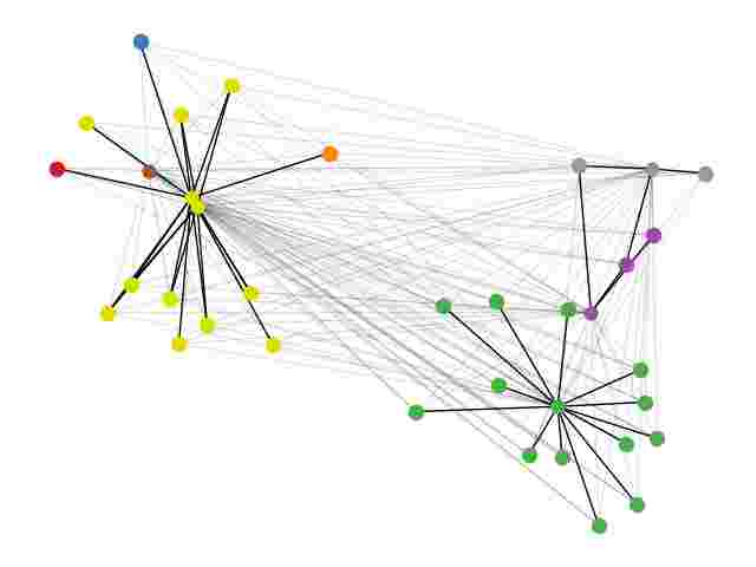}}
        \caption*{$G_2$}
    \end{subfigure}
    \caption{Usage-sense graph of Latin \textit{sacramentum} (left), subgraph for first time period $C_1$ (middle) and second time period $C_2$ (right).}\label{fig:graphsacrametum}
\end{figure*}

\subsection{Usage and sense sampling} 
\label{sec:use2}

For each target word, 30 usages from each of $C_1$ and $C_2$ containing $\geq 2$ tokens were randomly sampled, yielding a total of 60 usages per target word. 
The sense definitions were taken from the Latin portion of the Logeion online dictionary.\footnote{\url{https://logeion.uchicago.edu/}} Due to the challenge of finding qualified annotators, each word was assigned to one annotator, apart from \emph{virtus}, which was annotated by four annotators and used for inter-annotator agreement (Table \ref{tab:data}). The annotators could add comments to their annotations.
The senses and usages were presented in randomized order to the annotators.

\subsection{Edge sampling} 
\label{sec:edge2}

Procedure (ii) has an upper bound on the total number of annotated usage-sense pairs of $n \times k$ with $k$ senses for $n$ usages. The number of senses ranged between 2 and 7 with a usage sample size of $60$ which yielded a good number of annotation instances. Hence, no further optimization of the edge sampling procedure was carried out. Note though that a similar optimization as for procedure (i) would be possible by annotating the data incrementally or by randomly subsampling edges.

\subsection{Clustering} 
\label{sec:clustering2}

From the annotation, USGs where each usage is connected to each sense by one edge (see Figure \ref{fig:triplet2}) were obtained. Therefore there is a first-order path between each usage-sense pair and a second-order path between each usage-usage pair. Similarly to UUGs, we wanted to assign usages and senses into the same cluster if they received high judgments (3, 4) and into different clusters if they received low judgments (1, 2). We used the same clustering algorithm as for UUGs, defined in Section \ref{sec:clustering1}. In this way, usages end up in the same cluster if they have high judgments with the same senses. If there are contradictory judgments (e.g. a usage has high judgments with several senses), the clustering uses the global information to decide on the cluster assignment by choosing the one with the lowest loss. This can also lead to the collapsing of two sense descriptions into one cluster, e.g. for Latin \textit{sacramentum} in Figure \ref{fig:graphsacrametum}.

\begin{table}[t]
\setlength{\tabcolsep}{2pt}
\centering
\small
\begin{tabular}{ l c c c c c c c c c}            \toprule                           
\textbf{LGS}  &  $\mathbf{n}$  &  \textbf{N/V/A } &  $\mathbf{|U|}$  & \textbf{AN} & \textbf{ JUD } & \textbf{ AV } &  \textbf{SPR } &  \textbf{KRI } & \textbf{ LOSS } \\
\midrule                           
\textbf{EN} & 40 &  36/4/0  & 189 & 9 &  29k & 2  & .69 & .61 & .16 \\
\textbf{DE} & 48 &  32/14/2  & 178 & 8 &  37k & 2  & .59 & .53 & .12 \\
\textbf{SV} & 40 &  31/6/3  & 168 & 5 &  20k & 2 & .57 & .56 & .08 \\
\textbf{LA}  & 40 & 27/5/8 & 59 & 1 & 9k & 1 & .64 & .62 & .16  \\
\bottomrule                           
\end{tabular}   
\caption{Overview target words. LGS = language, $n$ = no. of target words, N/V/A = no. of nouns/verbs/adjectives, $|U|$ = avg. no. usages per word, AN = no. of annotators, JUD = total no. of judged usage pairs, AV = avg. no. of judgments per usage pair, SPR = weighted mean of pairwise Spearman in round 1, KRI = Krippendorff's alpha in round 1, LOSS = avg. of normalized clustering loss * 10.}
\label{tab:data}
\vspace{-9ex}
\end{table}

\section{Resource} 

A summary of the annotation outcome for each language can be found in Table \ref{tab:data}.
The final resource contains 40 words for EN/SV/LA, and 48 words for DE.\footnote{We release the data for all words including the ones which were excluded during the annotation process as described in Section \ref{sec:edge1}.} We report two annotation agreement measures: mean pairwise Spearman correlations \citep{bolboaca2006pearson} between annotator judgments and Krippendorff's alpha \cite{krippendorff2004content} for judgments' consensus, both reaching comparable scores to previous studies \citep{Erk13,Schlechtwegetal18,rodina2020rusemshift}. The clustering loss is the value of $L$ (Definition \ref{eq:loss}) divided by the maximum possible loss on the respective graph. It gives a measure of how well the graphs could be partitioned into clusters by the $L$ criterion.
In total, roughly 100,000 judgments were made by annotators. For EN/DE/SV $\approx $50\% of the usage pairs were annotated by more than one annotator, while for LA each target word but one was annotated by one annotator.

\begin{figure}[t]
    \centering
    \includegraphics[width=\linewidth]{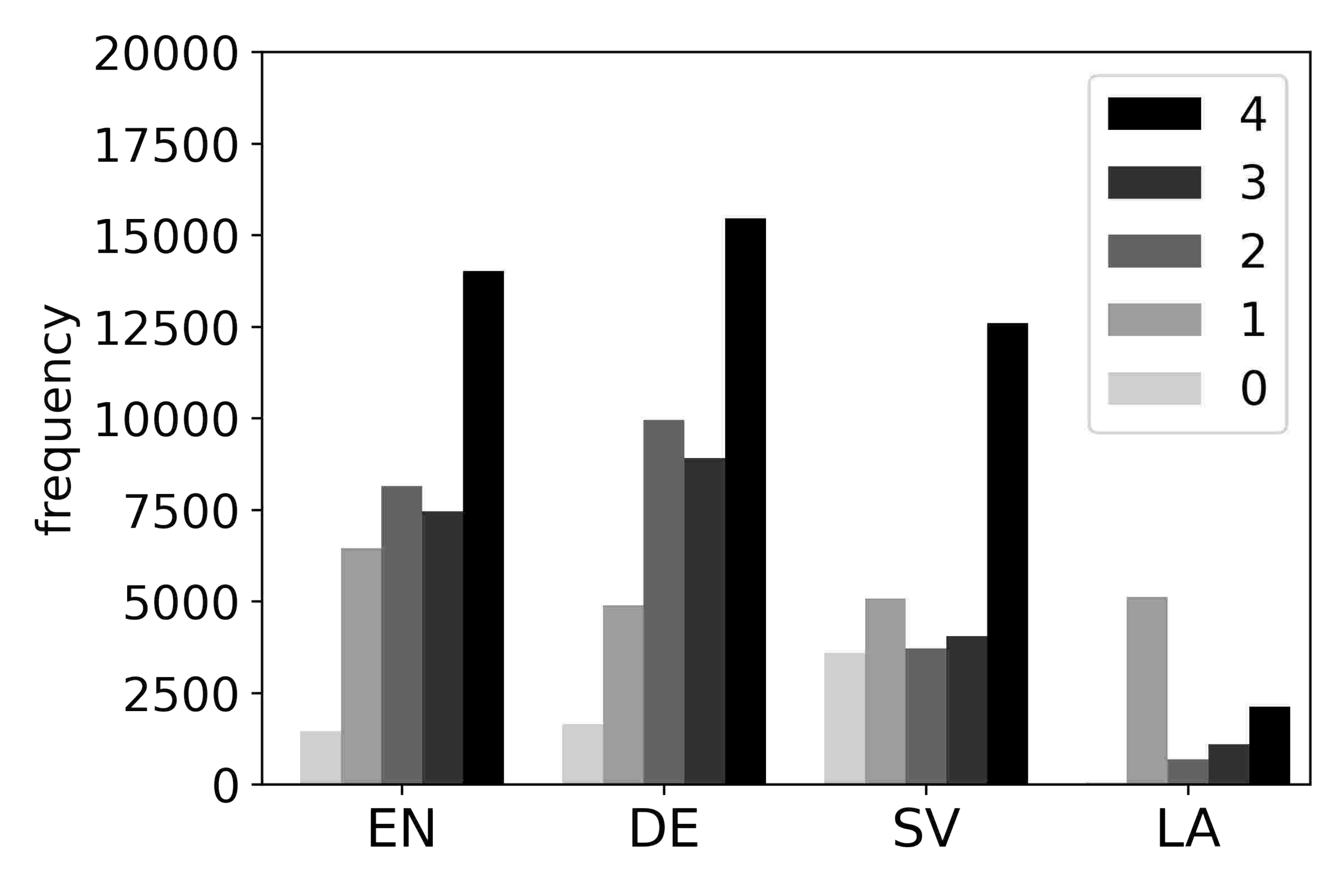}
    \caption{Judgment frequency per language.}
    \label{fig:judgment}
\vspace{-8ex}
\end{figure}

Figure \ref{fig:judgment} shows the frequencies of annotator judgments over the DURel scale by language. On the UUGs (EN/DE/SV) judgment `4' is most frequent followed either by judgment `2' (EN/DE) or `1' (SV). Least frequent are judgments of `0' (`Cannot decide'). Swedish has a considerably higher number of `0' judgments, presumably because of frequent OCR errors. On the USGs (LA) judgments of `1' are clearly most frequent, followed by `4'. This is because each usage is judged against each sense description which can often be unrelated. It can be seen that annotators make frequent use of the intermediate levels of the scale (`2', `3') and thus assign graded distinctions of word meaning.

\section{Analysis}

\subsection{Annotator disagreements}

\begin{figure}[t]
    \centering
    \includegraphics[width=\linewidth]{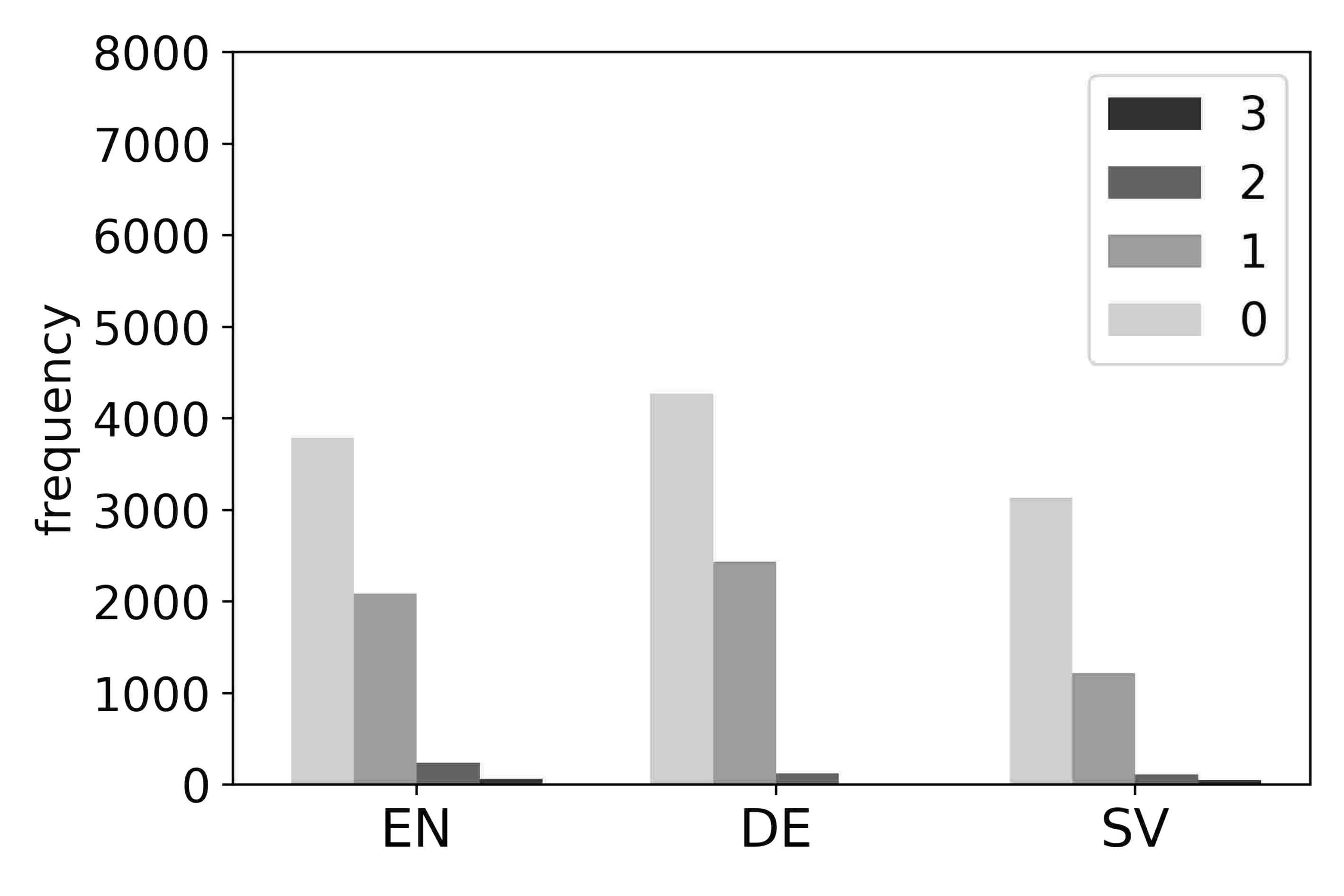}
    \caption{Disagreement frequency on edges with two annotations. Numbers in legend correspond to disagreements by points on the DURel scale.}
    \label{fig:disagreement}
\vspace{-8ex}
\end{figure}

Roughly half of all edges were annotated by only one annotator. In order to estimate the reliability of these annotations we report disagreement frequencies on all edges with two judgments as displayed in Figure     \ref{fig:disagreement}. Annotator pairs agree on 61--69\% of these edges across languages, while they disagree by one point on the scale on 27--34\%. Stronger disagreements are very rare with less than 5\%.

We further analyze annotator disagreements on a subset of words from the DWUG DE data set covering different POS (\textit{abbauen} (VB), \textit{abgebrüht} (ADJ), \textit{Knotenpunkt} (NN), \textit{Manschette} (NN), \textit{zersetzen} (VB)); we extract edges where at least one annotator pair diverges by at least two points on the DURel scale in Table \ref{tab:scales} (e.g. 1/3). We identify 5 sources of disagreement:
\begin{itemize}
 \setlength{\itemsep}{0pt}
\setlength{\parskip}{0pt}
\setlength{\parsep}{0pt}
    \item ambiguity
     \item meaning unfamiliarity
     \item misleading context
     \item unclear meaning abstraction level
     \item different intuitions on semantic proximity
\end{itemize}

Most cases of disagreements between annotators can be traced back to ambiguity or meaning unfamiliarity with one of the usages.
\begin{example}\label{ex:ambclear5}
 das war ein finsterer Herr mit dem harten Blick eines \textbf{abgebrühten} Schellfisches.\\
 `that was a sinister gentleman with the hard look of a \textbf{blanched/hard-nosed} haddock'
 \end{example}%
 \begin{example}\label{ex:ambcontext1}
 Darum hatte Calloway solche \textbf{Manschetten}, was?\\
 `That's why Calloway had \textbf{fear/cuffs/collars} like that, huh?'
 \end{example}%
 \begin{example}\label{ex:vague1}
 Vor allem Gregor Strasser war einer der braunen Halbgötter, bis er 1932 kurzerhand von Hitler \textbf{abgebaut} wurde.\\
 `Above all Gregor Strasser was one of the brown demigods until he was \textbf{destroyed?/deprived?} by Hitler in 1932.'
\end{example}
(\ref{ex:ambclear5}) is a case of ambiguity: \textit{abgebrüht} modifies an animal which could be ``blanched'' in the literal sense, but could also mean ``hard-nosed'' as the animal is further attributed with a ``hard glance''. Often ambiguity is also triggered by missing sentence context.
(\ref{ex:ambcontext1}) is a short sentence which gives little clues on the meaning of the target word. \textit{Manschetten} is at least ambiguous between a ``fear'', a ``cuff'' and a ``collar'' reading. 
In (\ref{ex:vague1}) \textit{abgebaut} occurs in an archaic sense which was only observed once in our data and is likely unfamiliar to annotators. The context and its other senses suggest a meaning like ``to destroy, to deprive'', but the exact meaning is unclear.
Further cases include usages with misleading context where a superficial reading or certain key words suggest a specific reading, while a deeper reading suggests another, and usages where the meaning of the target word could be described on various abstraction levels. There are also a few cases where the above categories do not apply, which may be due to (genuinely) different intuitions on semantic proximity.

\subsection{Robustness}
\begin{figure}[t]
    \centering
    \includegraphics[width=\linewidth]{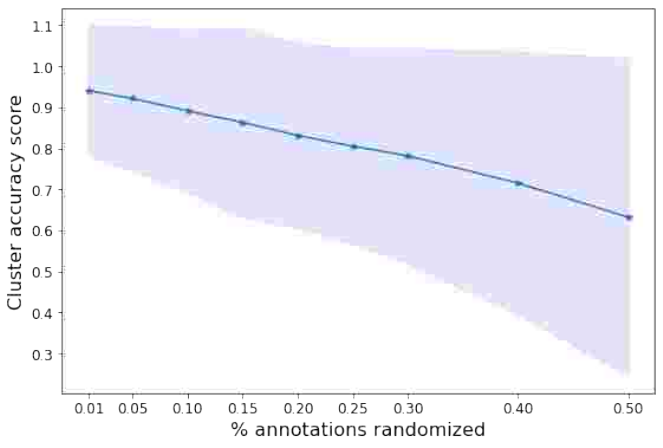}
    \caption{Mean cluster accuracies and CI (y axis) for increasing proportions of random annotations (x axis).}
    \label{fig:robustness}
\vspace{-8ex}
\end{figure}

To estimate if the clustering method is sensitive to spurious errors in the annotation procedure, we tested the robustness of our results to perturbations in the graphs' weights. We replaced existing annotations with random scores (i.e., changing scores only for existing annotation pairs), created new graphs, and clustered them. We then compared the similarity between the clusters in the original graphs, which we viewed as true labels, to those of the manipulated graphs using \textit{cluster accuracy}. This analysis, computed on English graphs (Figure \ref{fig:robustness}), demonstrates that the cluster structure of the graphs is robust under relatively high degree of random annotations: at an error rate of 25\% of the annotations, the manipulated graphs have cluster accuracy greater than 80\% on average.

\section{Conclusion}
We described the creation of the largest existing resource of word usage graphs that capture graded, contextualized word meaning for four languages, namely English, German, Swedish and Latin. We detailed the annotation procedure, including the sampling aimed to reduce annotation effort while keeping a high density in regions where annotators have difficulty judging relatedness. The usage graphs have been clustered and we openly release clusterings, visualizations and an analysis of the clustering results. 
This resource has been used for the SemEval 2020 task on unsupervised lexical semantic change detection, but its possibilities are much broader and range from the use of different clustering techniques, including soft-clustering, to the use as ground truth for diachronic word sense disambiguation or temporal classification of sentences. The corpora used and some aspects of the annotation procedure were different for Latin, and this was a necessary choice due to the lack of native speakers for this language and to the nature of the texts at our disposal. Offering a resource for Latin attests to the methodological and intellectual contribution of our work and we believe in the value of working on lexical semantic change for a historical language.

Future work entails annotating additional critical edges to allow for better understanding of robustness; how much annotation is needed for different kinds of words? Knowing that some words, e.g., single-sense concrete words, require less annotation allows us to spend more effort on abstract and highly polysemous words. We will also analyze the influence of edge sparsity and ambiguity on the clustering procedure and compare its output to other annotation strategies.

\section*{Acknowledgments}
The authors would like to thank Diana McCarthy for her valuable input to the genesis of this task. DS was supported by the Konrad Adenauer Foundation and the CRETA center funded by the German Ministry for Education and Research (BMBF) during the conduct of this study. The creation of the data was supported by the CRETA center and the CLARIN-D grant funded by the German Ministry for Education and Research (BMBF). This task has been funded in part by the project \textit{Towards Computational Lexical Semantic Change Detection} supported  by the Swedish Research Council (2019–2022; dnr 2018-01184), and \emph{Nationella språkbanken} (the Swedish National Language Bank) -- jointly funded  by  (2018--2024; dnr 2017-00626) and its 10 partner institutions, to NT. The Swedish list of potential change words were provided by the research group at the Department of Swedish, University of Gothenburg that work with the Contemporary Dictionary of the Swedish Academy. This work was supported by The Alan Turing Institute under the EPSRC grant EP/N510129/1. Additional thanks go to the annotators of our datasets, and an anonymous donor.

\newpage
\bibliography{dwug,Bibliography-general,bibliography-self}
\bibliographystyle{acl_natbib}

\newpage

\appendix

\section{Annotation pipeline example}
\label{sec:pipeline}

\begin{figure*}[t]
    \begin{subfigure}{0.33\textwidth}
\frame{        \includegraphics[width=\linewidth]{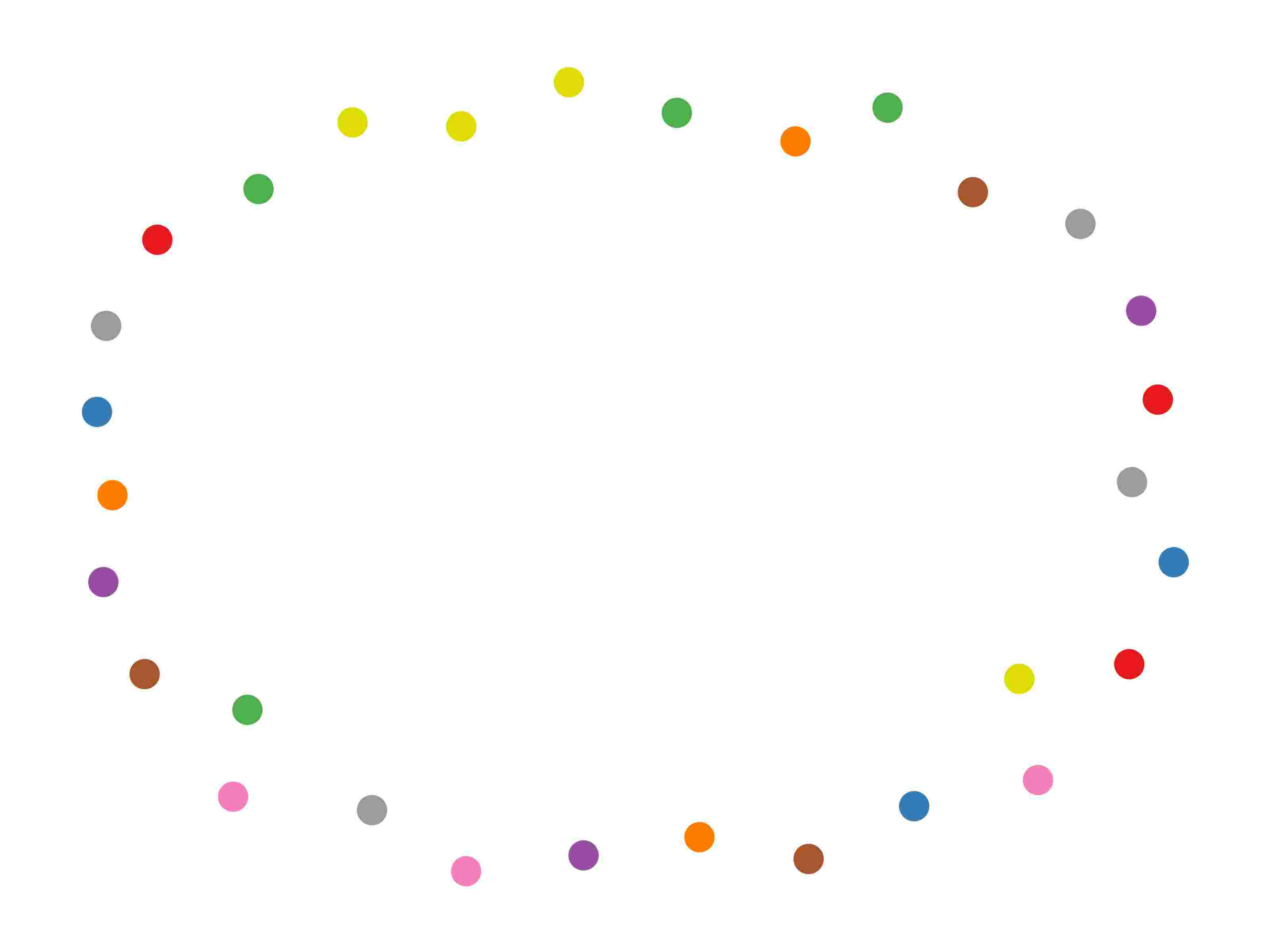}}
        \caption*{round 0}
    \end{subfigure}
    \begin{subfigure}{0.33\textwidth}
\frame {        \includegraphics[width=\linewidth]{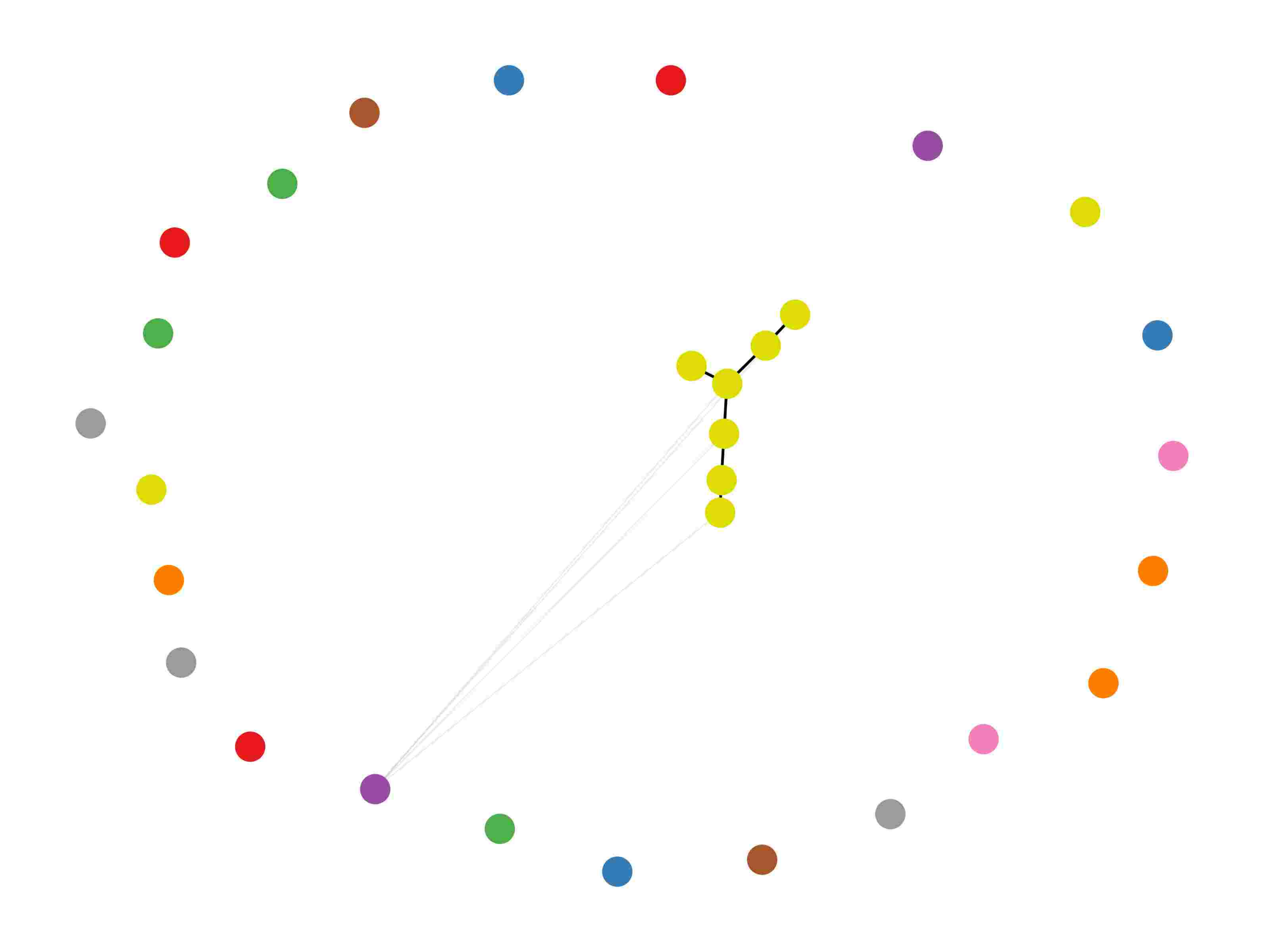}}
        \caption*{round 1}
    \end{subfigure}
    \begin{subfigure}{0.33\textwidth}
\frame {        \includegraphics[width=\linewidth]{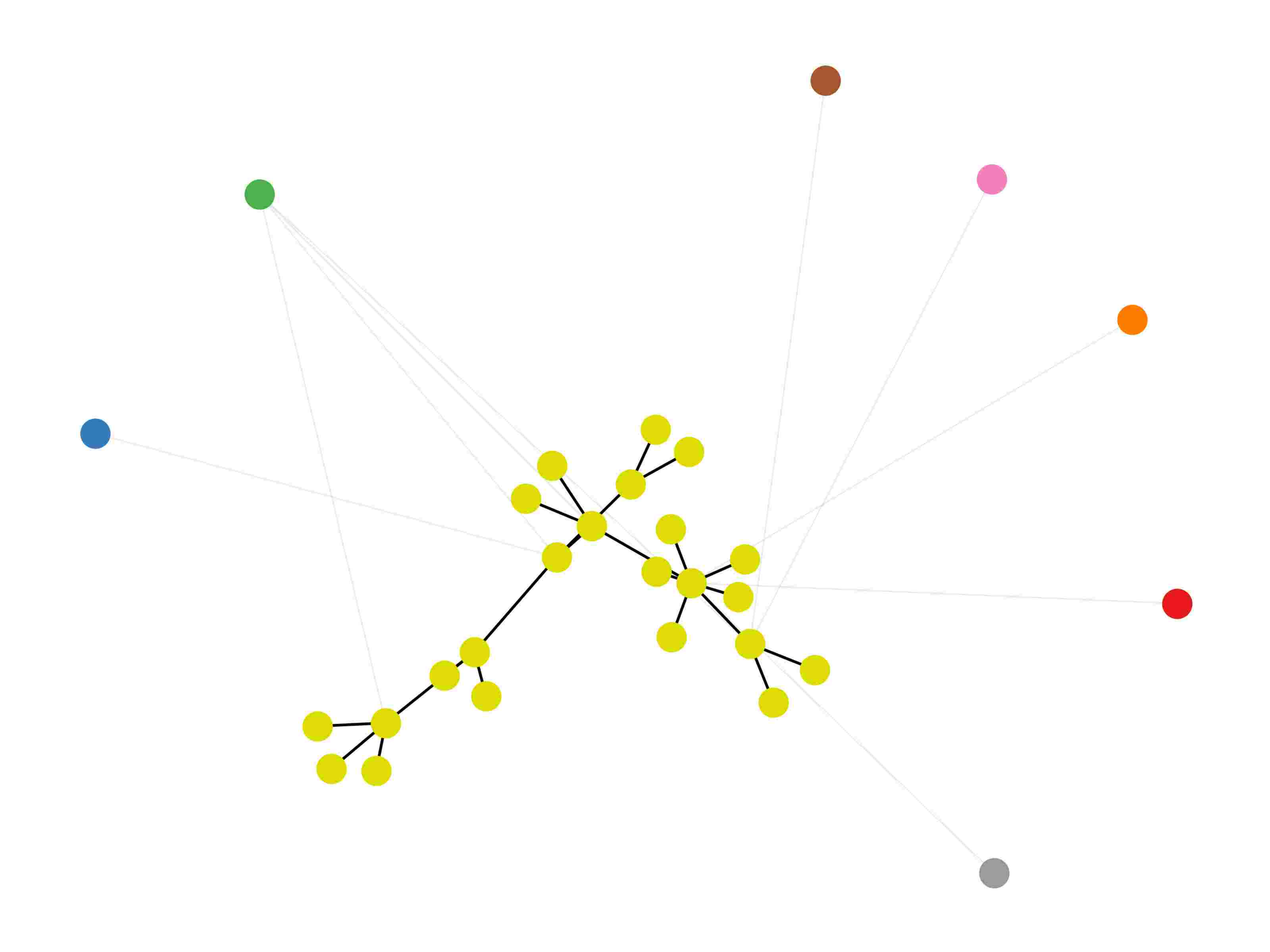}}
        \caption*{round 2}
    \end{subfigure}
    \begin{subfigure}{0.33\textwidth}
\frame {        \includegraphics[width=\linewidth]{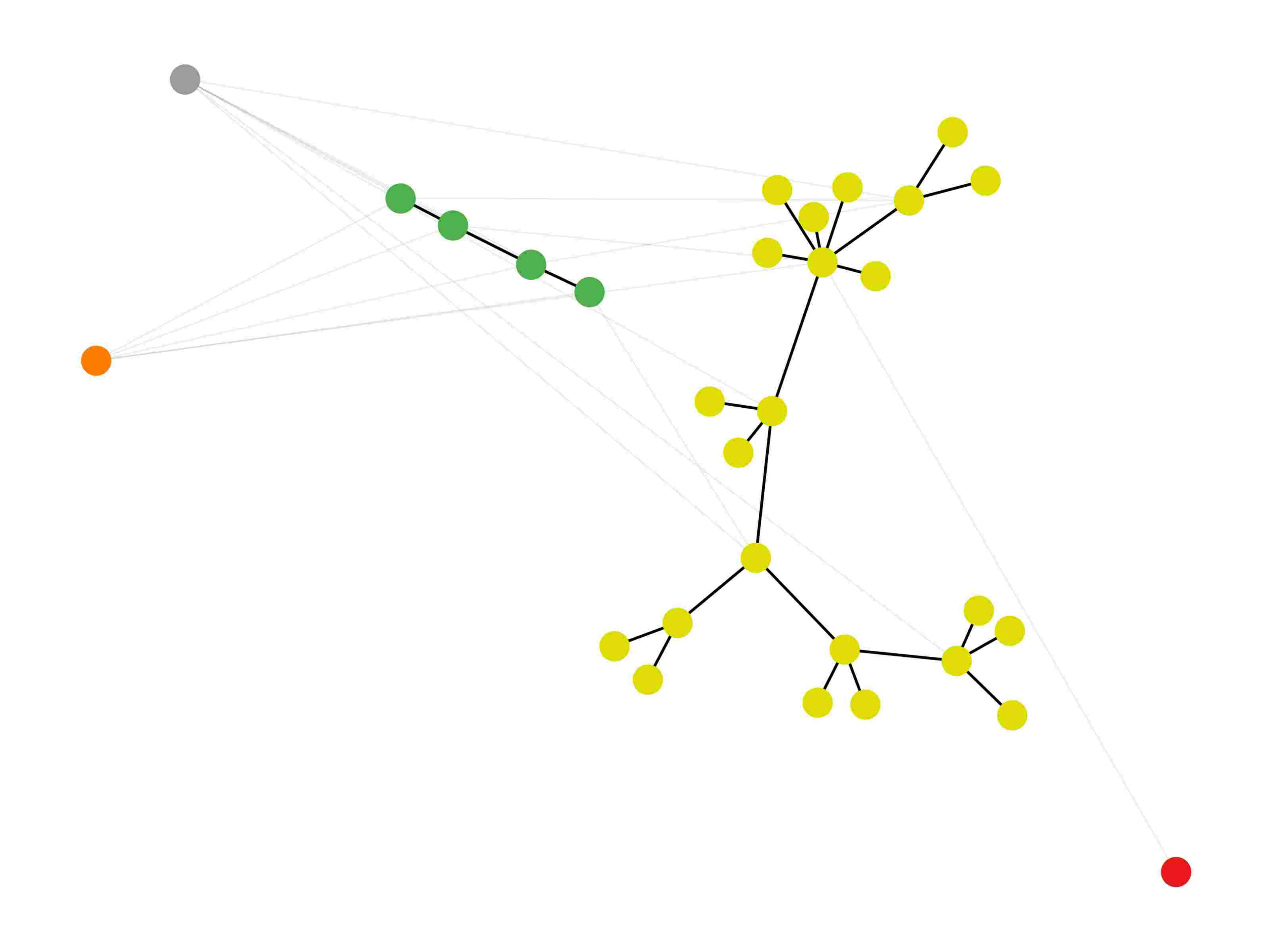}}
        \caption*{round 3}
    \end{subfigure}
    \begin{subfigure}{0.33\textwidth}
\frame {        \includegraphics[width=\linewidth]{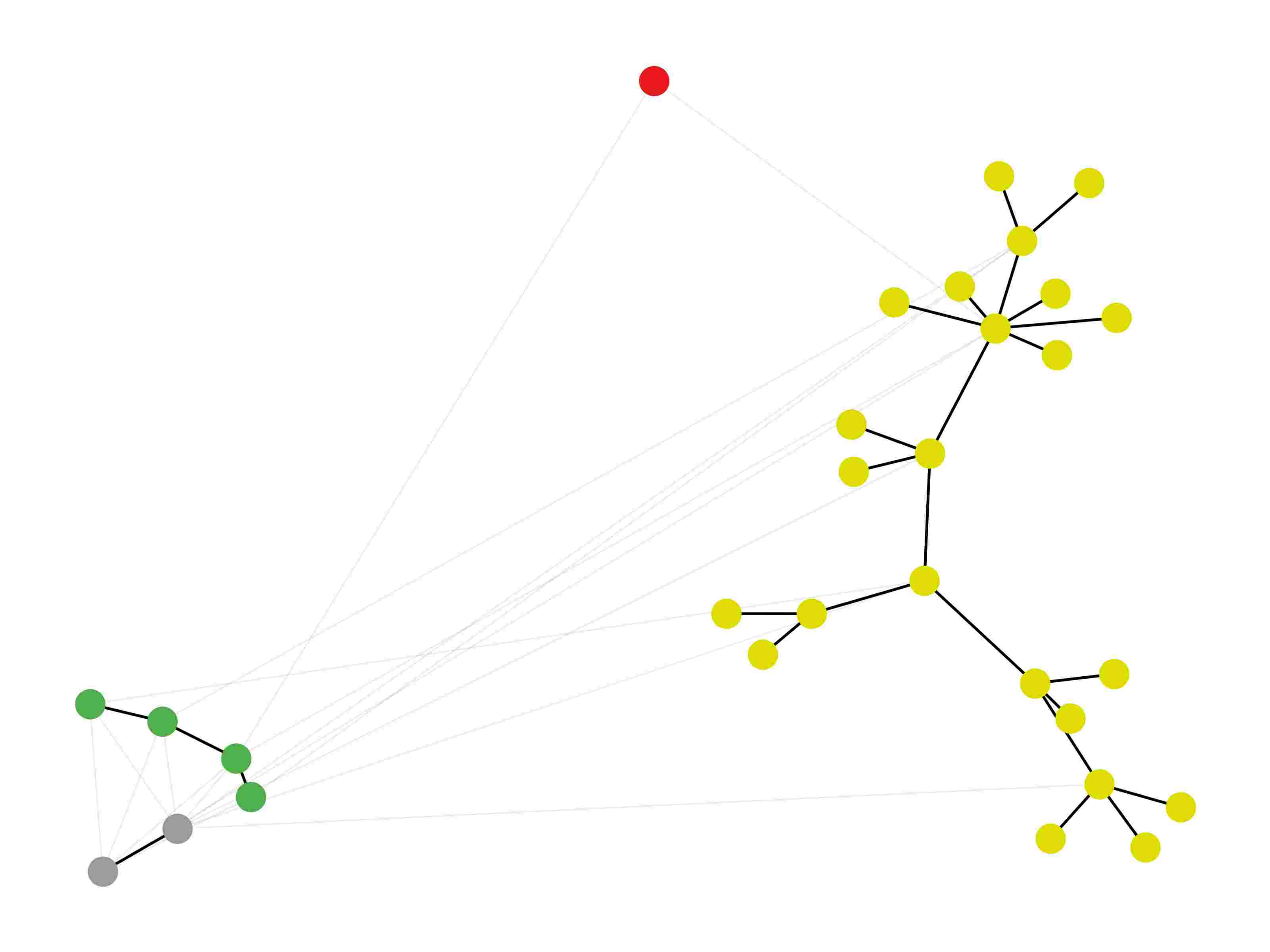}}
        \caption*{round 4}
    \end{subfigure}
    \begin{subfigure}{0.33\textwidth}
\frame {        \includegraphics[width=\linewidth]{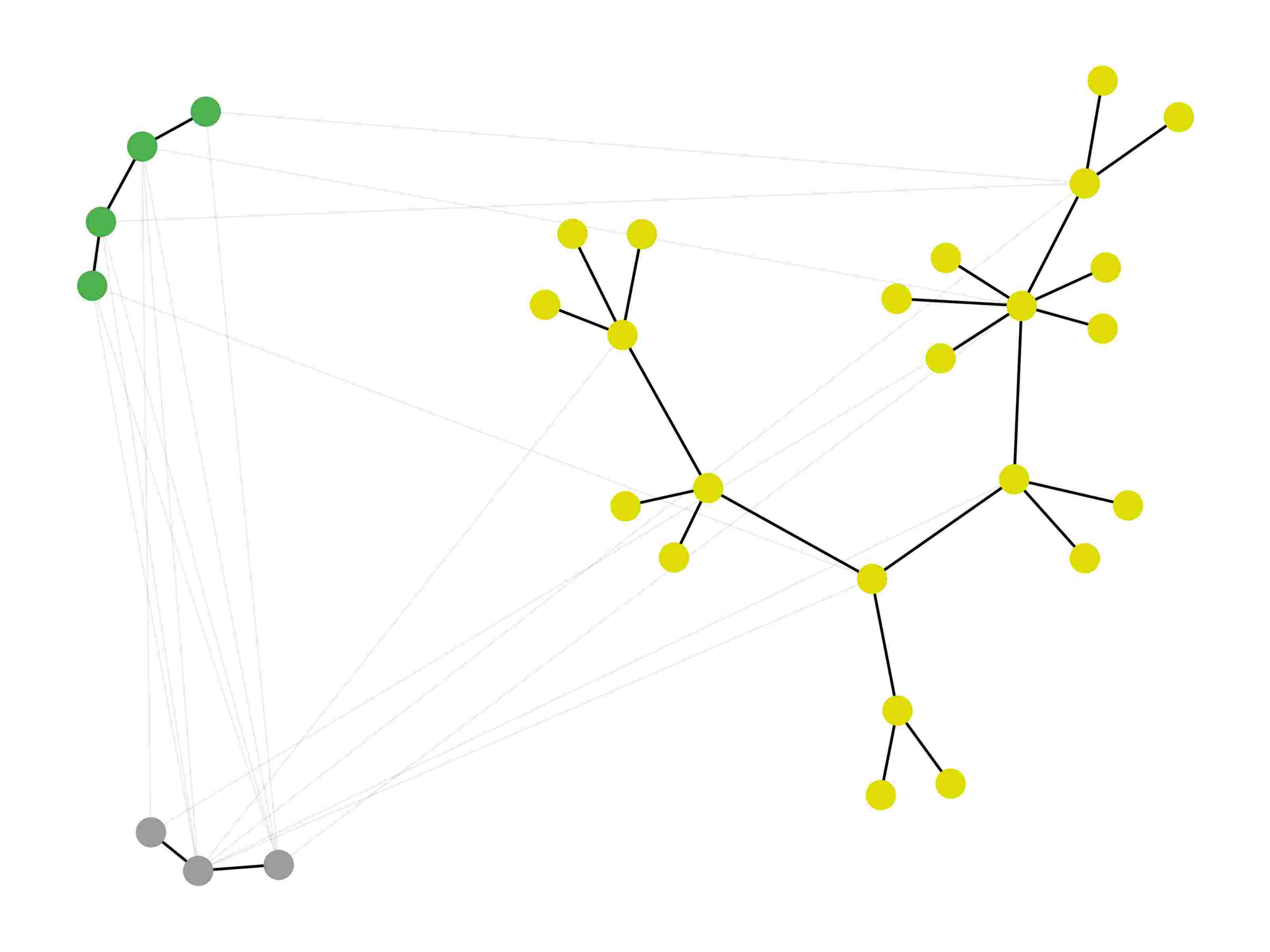}}
        \caption*{round 5}
    \end{subfigure}
    \caption{Simulated example of annotation pipeline.}\label{fig:pipeline}
\end{figure*}

Figure \ref{fig:pipeline} shows an example of our annotation pipeline. As the annotation proceeds through the rounds, the graph becomes more populated and the true cluster structure is found. In round 1 one multi-cluster is found. Hence, all remaining usages are compared with this cluster in round 2 by the combination step. In rounds 3 and 4 the exploration step discovers more clusters not found in the rounds before.

\end{document}